\documentclass[11pt,a4paper]{article}
\usepackage[hyperref]{acl2020}
\usepackage{times}
\usepackage{latexsym}

\usepackage{microtype}

\usepackage{kotex-logo}
\usepackage{multirow}
\usepackage{graphicx}
\usepackage{bm}

\title{Stable Style Transformer: Delete and Generate Approach with Encoder-Decoder for Text Style Transfer}

\author{First Author \\
  Affiliation / Address line 1 \\
  Affiliation / Address line 2 \\
  Affiliation / Address line 3 \\
  \texttt{email@domain} \\\And
  Second Author \\
  Affiliation / Address line 1 \\
  Affiliation / Address line 2 \\
  Affiliation / Address line 3 \\
  \texttt{email@domain} \\}

\date{}

\begin{document}
\maketitle
\begin{abstract}
Text style transfer is the task that generates a sentence by preserving the content of the input sentence and transferring the style. Most existing studies are progressing on non-parallel datasets because parallel datasets are limited and hard to construct. In this work, we introduce a method that follows two stages in non-parallel datasets. The first stage is to delete attribute markers of a sentence directly through a classifier. The second stage is to generate a transferred sentence by combining the content tokens and the target style.  We experiment on two benchmark datasets and evaluate context, style, fluency, and semantic. It is difficult to select the best system using only these automatic metrics, but it is possible to select stable systems. We consider only robust systems in all automatic evaluation metrics to be the minimum conditions that can be used in real applications. Many previous systems are difficult to use in certain situations because performance is significantly lower in several evaluation metrics. However, our system is stable in all automatic evaluation metrics and has results comparable to other models. Also, we compare the performance results of our system and the unstable system through human evaluation. Our code and data are available at the link~\footnote{https://github.com/rungjoo/Stable-Style-Transformer}.
\end{abstract}

\section{Introduction}
Text style transfer is a task that generates a sentence while preserving the content in a given sentence but changing the source style. The style of the sentence refers to a predefined class (e.g. sentiment, formality, tense) and the content refers to the rest of the sentence except for the style. Lack of parallel data makes text style transfer task difficult. This problem cannot be solved by supervised learning because there are no right sentences.

One previous method~\cite{hu2017toward,shen2017style,fu2018style,prabhumoye_etal_2018_style,logeswaran2018content} of text style transfer is to learn latent representations to separate style and content from sentences. First, these approaches try adversarial training to learn a disentangled latent representation of the content and style. Secondly, a transferred sentence is generated from the decoder by combining the disentangled latent representation and the target style. However, the experimental results of~\cite{lample2018multipleattribute} report that disentangled latent representation through adversarial training is hard to get and not necessary. Also, adversarial training is not effective to encode a sentence of various lengths into a vector representation of fixed length. Other methods of text style transfer do not depend on disentanglement. \citet{dai_etal_2019_style,lample2018multipleattribute,ijcai2019-711} do not attempt to find the disentangled latent representation in the sentence. Therefore, sentences with different styles are mapped to the same space. \citet{xu_etal_2018_unpaired,li_etal_2018_delete,sudhakar-etal-2019-transforming,ijcai2019-732} neutralize sentences by deleting style-dependent attribute markers. Remaining tokens resulting from the deletion of attribute markers are style independent, and then the content tokens and a style attribute are combined to generate the transferred sentence.

We propose an approach with two stages using Delete and Generate without adversarial training for disentanglement. (1) Attribute markers of a sentence are extracted by using a pre-trained classifier as a Delete model. Our method is model-agnostic and is not affected by the design of the classifier. Attribute markers found in a sentence are deleted. (2) A transferred sentence is generated by combining the target attribute and the content tokens after stage-1. The Generate model consists of an encoder and decoder with the Transformer structure.

In the method of deleting attribute markers, \citet{li_etal_2018_delete} deletes attribute markers via a statistical manner using a frequency ratio and \citet{sudhakar-etal-2019-transforming,xu_etal_2018_unpaired} delete attribute markers using attention weights of a classifier. \citet{ijcai2019-732} deletes attribute markers by fusion of the frequency ratio and the attention weights. We introduce an intuitive delete method that uses a change in classifier probability. If a change in classifier probability is significant when limiting certain tokens in a sentence, the token is considered an attribute marker. Our method does not need to build attribute dictionaries or define attention weights like previous methods and easily control the trade-off between content and style.

We test our methods on two text style transfer datasets: sentiment of Yelp reviews and Amazon reviews. Evaluation metrics are conducted in terms of content, fluency, style accuracy, and semantic. The content and style accuracy are measured similarly to previous studies. Fluency is measured in two ways: general-fluency using pre-trained GPT-2~\cite{radford2019language} and data-fluency using finetuned GPT-1~\cite{Radford2018ImprovingLU}. Semantic is newly evaluated using BERTscore~\cite{zhang*2020bertscore} in this paper. The goal of BERTscore is to evaluate semantic equivalence between two sentences. In this paper, we use a pre-trained model GPT and BERT~\cite{devlin_etal_2019_bert} that perform well in natural language processing/generation to evaluate transferred sentences with various automatic evaluations. Since automatic evaluations are not perfect evaluations of generated sentences, it is hard to know which system is the best, but we can determine which system has a problem. Comparative models are unstable in some evaluation metrics. But our proposed model has stable results for all automatic evaluations and is called SST (Stable Style Transformer). In addition, we first observe a point that can enhance the style controlling ability by generating sentences through latent space walking in the vector space of the style attribute token.

\section{Related Work}
One line of text style transfer research~\cite{shen2017style,fu2018style,hu2017toward,prabhumoye_etal_2018_style,logeswaran2018content} is to separate content and style from sentences through disentangled learning. \citet{hu2017toward} uses the VAE model to derive the disentanglement of the content between the generated sentence and the original sentence through KL loss. \citet{shen2017style} introduce the aligned auto-encoder and the cross aligned auto-encoder using learning discriminators. \citet{fu2018style} propose a multi-decoder and StyleEmbedding model. The multi-decoder model has decoders for each style, and the style embedding model uses only one decoder by inserting style embedding into the decoder. The methods of~\citet{prabhumoye_etal_2018_style,logeswaran2018content} used back-translation to learn latent representations.

The second line of text style transfer research is not to rely on learning for latent representation. The first approach~\cite{xu_etal_2018_unpaired,li_etal_2018_delete,sudhakar-etal-2019-transforming,ijcai2019-732} is to find and delete tokens called attribute markers that are highly related to style. \citet{li_etal_2018_delete} uses the delete method of attribute markers as a statistical method based on frequency ratio, and \citet{sudhakar-etal-2019-transforming,xu_etal_2018_unpaired} use the attention scores of the Transformer classifier and LSTM classifier, respectively. \citet{ijcai2019-732} deletes attribute markers by fusion of the frequency ratio and attention scores. The second approach~\cite{dai_etal_2019_style,lample2018multipleattribute,ijcai2019-711} does not attempt to control content and style separately. Therefore, sentences with different styles are encoded to gather in the same latent representation space. \citet{dai_etal_2019_style,lample2018multipleattribute} are based on the learning method using cycle reconstruction loss. \citet{lample2018multipleattribute} reported that disentanglement is not easy and that latent representations learned through adversarial training are unnecessary because learned latent representations depend on style. Unlike the previous models, \cite{ijcai2019-711} learns dual models in two directions: style1 (e.g. negative) to style2 (e.g. positive) and style2 (e.g. positive) to style1 (e.g. negative) by reinforcement learning. 
\begin{figure*}[!htb]
    \centering 
    \includegraphics[width=2.0\columnwidth]{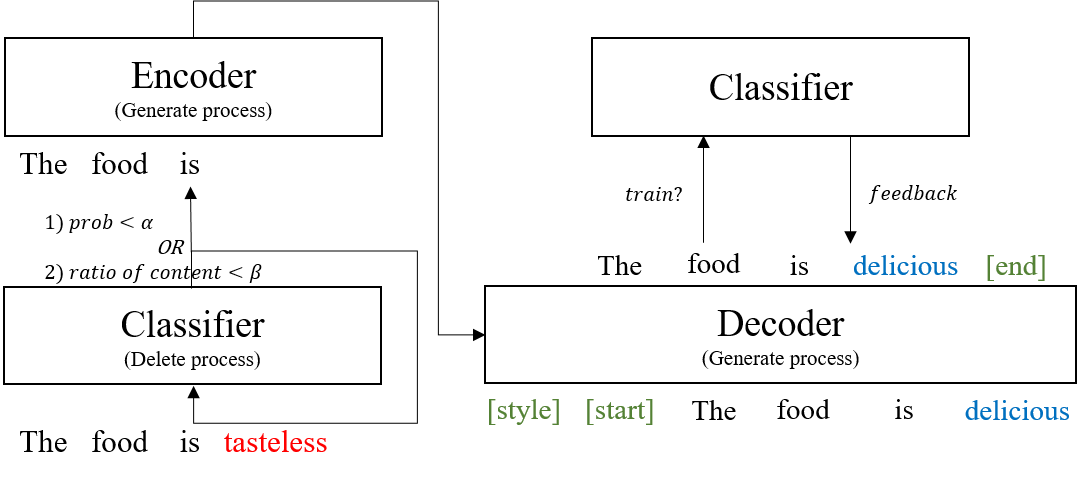}
    \caption{The proposed model framework consists of Delete and Generate process. Delete process is a method using a pre-trained classifier, and the Generate process consists of an encoder and a decoder. In the training time, our model receives feedback from the classifier's probability of the generated sentence.}
    \label{fig:model}
\end{figure*}

In language model research, the RNN-based language model is weak in long dependency. Therefore, the recent study of text style transfer~\cite{dai_etal_2019_style,sudhakar-etal-2019-transforming,ijcai2019-732} has been conducted with Transformer~\cite{NIPS2017-7181} which is known to have good performance in language modeling. \citet{dai_etal_2019_style} uses a method of using the encoder and the decoder of the Transformer, and \citet{sudhakar-etal-2019-transforming} uses a method of fine-tuning the decoder to the style transfer datasets with the pre-trained GPT-1 as an initial state. \citet{ijcai2019-732} solved the problem of text style transfer in a similar way to Text Infilling or Cloze by presenting Attribute Conditional Masked Language Model (AC-MLM) using pre-trained BERT.

In this paper, we chose the first approach (Delete and Generate) that does not rely on latent representations in the second research line, referring to the results of \citet{lample2018multipleattribute}. Our system has a Transformer encoder and decoder because the style transfer task is given input text. If the system uses only a decoder such as \citet{sudhakar-etal-2019-transforming}, there is a disadvantage that it cannot include bidirectional encoding of the content token. Or, if only bidirectional encoders are used, such as AC-MLM, the position and length of the masking tokens to be filled in a sentence is not flexible.

\section{Approach}
In this section, we introduce our proposed method. The style transfer problem definition is described in Section~\ref{problem-Statement}. An overview of the model is shown in Section~\ref{model-overivew}. The proposed generation process is introduced in Sections~\ref{delete-process} and \ref{generate-process}. The learning mechanism is described in Section~\ref{training}.

\subsection{Problem Statement}
\label{problem-Statement}
Given a dataset consist of sentence and label: $D=\{ (\bm{x}_1, s_1), \cdots, (\bm{x}_N, s_N) \}$ where $\bm{x}_i$ is a sentence and $s_i$ is a style attribute (e.g. sentiment) and N is the number of sentences in the dataset. Our goal is to train the model to generate a sentence $\bm{y}_i$ with a different style while preserving the content of the sentence $\bm{x}_i$.  For example, if $\bm{x}_i$ is \textit{"The food is salty and tasteless"} and $s_i$ is \textit{"negative"} attribute, then $\bm{y}_i$ is generated to mean \textit{"The food is not salty and delicious"} which has a \textit{"positive"} attribute. However the dataset is non-parallel, so the model cannot access $\bm{y}_i$ aligned with $\bm{x}_i$.

\subsection{Model Overview}
\label{model-overivew}
Our approach consists of two stages: Delete and Generate framework in Fig.~\ref{fig:model}. The first stage is the Delete process with a pre-trained style classifier. The pre-trained style classifier finds and deletes tokens that contain a lot of style attributes. The second stage is encoding the content tokens and combine them with a target style to generate a sentence. Both the encoder and the decoder have the Transformer structure, which is better than RNN and robust to long dependency.

\subsection{Stage-1: Delete process}
\label{delete-process}
The stage-1 is the process of finding and deleting tokens for a given sentence and style attribute. In the previous studies, the strategies of deleting attribute markers are the frequency-ratio method and the classifier's attention score (or fusion of both). However, the frequency ratio method requires a pre-built vocabulary for the training dataset and it is difficult to understand contextual information. The attention score method has a limitation on the structure of the classifier, because it must learn the style classifier using self-attention regardless of accuracy. It is also unclear whether the attention score is directly proportional to the attribute.

We propose a novel method of removing attribute markers using a pre-trained classifier without a pre-built dictionary and attention scores. Our method is a model-agnostic method and it is more intuitive to find attribute markers than the previous method. Given an input sentence $\bm{x}$, the style probability follows:
\begin{equation}
\label{prob1}
p_{\bm{x}} = p_{\theta_C}(s|\bm{x})
\end{equation}
where $p$ is a probability predicted by the classifier and $s$ is style label. If we delete token $t_i$ from the sentence $\bm{x}$, the style probability changes as follows:
\begin{equation}
\label{prob2}
p_{\bm{x}, t_i} = p_{\theta_C}(s|\bm{x}, t_i)
\end{equation}
where $\bm{x}=(t_1, t_2, \cdots, t_n)$ and $n$ is the number of tokens in $\bm{x}$. The probability difference between Eq. \ref{prob1} and Eq. \ref{prob2} is defined as $\textit{Important}$ $\textit{Score(IS)}$ of the token($t_i$):
\begin{equation}
\label{prob_diff}
\textit{IS}^{k}_{t_i}=p_{\bm{x}^{k}}-p_{\bm{x}^{k}, t_i}
\end{equation}
where $\bm{x}^{k}$ is the remained tokens after $k$ tokens are deleted. The value of $\textit{IS}^{k}_{t_i}$ determines how much the token ${t_i}$ affects the style classifier. The token is deleted in order of the largest $\textit{IS}$, and the Delete process ends if only one of the following two conditions: (1) $p_{\bm{x}^k}$ is less than $\alpha$, or (2) the ratio of content tokens is less than $\beta$. $\alpha$ is a hyperparameter that determines that a sentence no longer has a source style attribute. $\beta$ is a hyperparameter that determines how much of the content is preserved. The two hyperparameters make it easy to control the trade-off of content and style, and the experimental results are explained in Section~\ref{sec:trade-off}.

\subsection{Stage-2: Generate process}
\label{generate-process}
Our model generates a transferred sentence with the encoder and the decoder of the Transformer.

\subsubsection{Encoder}
All content tokens given as a result of Delete process are input to a bidirectional self-attention the Transformer encoder. Explicitly, the Transformer encoder maps content tokens $\bm{x}^c=(t_1,\cdots,t_m)$ to the continuous representation $\bm{z}=(z_1,\cdots,z_m)$ as follow:
\begin{equation}
\label{encoder}
(z_1,\cdots,z_m) = Encoder(t_1,\cdots,t_m;\theta_E)
\end{equation}

\subsubsection{Decoder}
In order to generate a sentence with the desired style, two special tokens, $\textit{style}$ and $\textit{start}$, are initially input to the decoder in Fig.~\ref{fig:model}. The position of special tokens is always fixed in front, so we did not add positional embedding. We use teacher-forcing at training time and no teacher-forcing at test time to generate sentences. If the generated token is the special token $\textit{end}$, the Generate process ends. The decoder auto-regressively predicts the conditional probability of the next step token as follows:
\begin{equation}
\label{decoder}
\textit{softmax}(\bm{y}_j) = p_{\theta_D}(t'_j|t'_1,\cdots,t'_{j-1}, \tilde{s}, \bm{z})
\end{equation}
where $\bm{y}_j$ is the logit vector of the decoder, $\tilde{s}$ is a desired style and $t'_j$ is the predicted token in $j$ step.

\subsection{Training}
\label{training}
Since we only have non-parallel datasets, we can't do supervised learning about transferred sentences. Therefore, we train SST to minimize two losses according to style conditions: $s$ (source style) or $\hat{s}$ (target style).

\subsubsection{Reconstruction loss}
SST reconstructs the original sentence $\bm{x}$ conditioned on $\bm{x}^c$ and source style $s$. Reconstruction loss follows the equation:
\begin{equation}
\label{reconstruction_loss}
\mathcal{L}_{rec} = -log\ p_{\theta_E, \theta_G}(\bm{x}|\bm{x}^c, s)
\end{equation}
In non-parallel datasets, the reconstruction loss cannot be calculated if the style of the generated sentence is $\hat{s}$.

\subsubsection{Style loss}
If the model is only trained with reconstruction loss, the decoder will not see how to transform the style. Therefore, a discrepancy occurs between training time and test time. To learn how to generate sentence $\hat{\bm{x}}$ with a target style $\hat{s}$, we introduce style loss as follows:
\begin{equation}
\label{style_loss}
\mathcal{L}_{style} = -log\ p_{\theta_C}(\hat{\bm{x}}=\hat{s}|\bm{x}^c, \hat{s})
\end{equation}
Style loss is measured by a pre-trained classifier to determine whether the transferred sentence has a $\hat{s}$. Since the generated sentence is a discrete space, we utilize soft-embedding of predicted tokens to optimize through style loss. When the SST is trained, the parameters of the classifier are not finetuned.

\subsection{Model Details}
The Transformer encoder and decoder consist of 3 layers, and each layer has 4 heads. The style classifier consists of 5 convolution filters based on \citet{kim_2014_convolutional}. Text is tokenized using Byte-Pair-Encoding, and (word, style, position) embeddings are 256-dimensional vectors. In the Delete process, $(\alpha, \beta)$ is $(0.7, 0.5)$ at training time and observes the trade-off of content and style by changing parameters during test time.

\section{Experiments}
\subsection{Dataset}
In this paper, we test our model on two datasets, \textrm{YELP} and \textrm{AMAZON}, which are provided in \citet{li_etal_2018_delete}. The Yelp dataset is for business reviews, and the Amazon dataset is product reviews.
Both datasets are labeled negative and positive and statistics are shown in Table~\ref{Tab:data}.

\subsection{Human References}
Human references are used to measure human-BLEU and BERTscore. We used 2 Yelp human references and 1 Amazon human reference. \textbf{Yelp}: \citet{li_etal_2018_delete} provides 1 human reference and 3 additional human references in \citet{ijcai2019-711}. We used 2 human references, one from \citet{li_etal_2018_delete} and one (the best performance in automatic evaluation) from \citet{ijcai2019-711}, to increase reliability. \textbf{Amazon}: We used the human reference provided by \citet{li_etal_2018_delete}.

\begin{table}[!t]
    \centering
    \resizebox{0.8\columnwidth}{!}{
        \begin{tabular}{|c|c|c|c|c|}
        \hline
        Dataset                 & Style    & Train & Dev  & Test \\ \hline
        \multirow{2}{*}{Yelp}   & Positive & 270K  & 2000 & 500  \\ \cline{2-5} 
                                & Negative & 180K  & 2000 & 500  \\ \hline
        \multirow{2}{*}{Amazon} & Positive & 277K  & 985  & 500  \\ \cline{2-5} 
                                & Negative & 278K  & 1015 & 500  \\ \hline
        \end{tabular}
    }
    \caption{(Sentiment) Dataset statistics}
    \label{Tab:data}
\end{table}

\subsection{Previous Method}
We compare the previous models with three approaches. The first comparisons are CrossAligned~\cite{shen2017style}, [StyleEmbedding, multi-decoder]~\cite{fu2018style}, and BackTranslation~\cite{prabhumoye_etal_2018_style}, which attempt to separate content and style through latent representation learning. The second comparisons are [DeleteOnly, DeleteAndRetrieve]~ \cite{li_etal_2018_delete}, UnpariedRL~\cite{xu_etal_2018_unpaired} and [B-GST, G-GST]~\cite{sudhakar-etal-2019-transforming}, which delete attribute markers and then generate the sentence. [TemplateBased, RetrieveOnly]~\cite{li_etal_2018_delete} return the target sentence through retrieve without generating. The final comparison is DualRL~\cite{ijcai2019-711}, which does not distinguish between content and style.

\subsection{Automatic Evaluation}
We evaluated the systems in 4 ways and results are shown in Table~\ref{Tab:Yelp} and \ref{Tab:Amazon}.

\subsubsection{Content}
Content preserving intensity is measured by G-BLEU, the geometric mean of self-BLEU and human-BLEU, as in previous works. A high BLEU score indicates that the model is good at content preservation.

In the Yelp dataset, RetireveOnly and BackTranslation are considered unstable models because G-BELU score is too low compared to other systems. In the Amazon datasets, CrossAligned and RetrieveOnly are too low compared to other systems.

\subsubsection{Attribute}
Most style transfer studies measure style accuracy using a classifier. We also evaluate style accuracy with a classifier (note that this is different from the one used in training).

In the Yelp dataset, StyleEmbedding, multi-decoder, and UnpairedRL have quite a low accuracy. In the Amazon datasets, StyleEmbedding, DeleteOnly, and DeleteAndRetrieve are unstable in style transfer. 

\subsubsection{Fluency}
Fluency is considered the perplexity of the transferred sentence. We use GPT-1 and GPT-2, which is known to perform well as a language model. General-fluency (g-PPL) is measured using pre-trained GPT-2 and data-fluency (d-PPL) is measured using GPT-1 (instead of GPT-2 due to GPU memory) finetuned to the dataset. General-Fluency is a general view because the language model is not fitted to the data, and data-fluency is an evaluation metric in terms of the specific data of style transfer tasks. The total-fluency (t-PPL) is the geometric mean of d-PPL and g-PPL, and lower values indicate better fluency.

In the Yelp dataset, TemplateBased is unstable because t-PPL is much larger than other systems. In the Amazon dataset, it is determined that the fluency of B-GST and G-GST is unstable.

\subsubsection{Semantic}
Semantic is measured using BERTscore. Unlike BLEU and ROUGE, BERTscore is an evaluation metric defined in continuous space. Pre-trained model is used to calculate cosine similarity by extracting the contextual token embeddings from a human reference and a transferred sentence. BERTscore solves the limitations of previous metrics and measures a better correlation between the reference and the candidate. The original BERTscore ranged from 0 to 1, but we rescale it from 0 to 100 to clearly see the difference.

We set the unstable threshold as a margin point lower than the mean of all systems. The margin point is a gap between an average and a lower bound with $95\%$ confidence considering all systems as the gaussian distribution in the BERTscore evaluation. CrossAligned, multi-decoder, RetrieveOnly, and BackTranslation have limitations on Yelp datasets. CrossAligned, multi-decoder, and RetrieveOnly have limitations on Amazon datasets.

\paragraph{SST}: For comparison with other systems, we select the $\alpha$ and $\beta$ of the appropriate trade-off points for style transfer and content preservation. When experimenting with the Yelp datasets, SST model is evaluated in two cases where $(\alpha, \beta)$ is (0.7, 0.5) and (0.7, 0.75). SST{\footnotesize{(0.7, 0.5)}} changes styles better with style accuracy of 79.5$\%$, but SST{\footnotesize{(0.7, 0.75)}} has better performance on other metrics. In the Amazon datasets, SST model is evaluated when $(\alpha, \beta)$ is (0.6, 0.5). The effects of $\alpha$ and $\beta$ are discussed in detail in Section~\ref{sec:trade-off}.

\begin{table*}[!h]
\centering
\resizebox{2.0\columnwidth}{!}{
\begin{tabular}{c|c|c|c|c|c|c|c|c|}
\cline{2-9}
                                                                                   & \multicolumn{3}{c|}{Content} & Attribute                            & \multicolumn{3}{c|}{Fluency}                               & Semantic  \\ \hline
\multicolumn{1}{|c|}{Model}                                                        & s-BLEU   & h-BLEU  & G-BLEU  & Classifier($\%$)                           & d-PPL   & g-PPL  & t-PPL                                   & BERTscore \\ \hline\hline
\multicolumn{1}{|c|}{SST \scriptsize{(0.7, 0.5)}}                                               & 39.05    & 10.85   & 20.58   & \textbf{79.5}                                 & 185.26  & 321.84 & 244.18                                  & 88.72     \\ \hline
\multicolumn{1}{|c|}{\begin{tabular}[c]{@{}c@{}}SST \scriptsize{(0.7, 0.75)}\end{tabular}} & 49.09     & 12.66   & \textbf{24.93}    & 70.4                                 & 197.82  & 295.9 & \textbf{241.94}                                  & \textbf{90.65}     \\ \hline\hline
\multicolumn{1}{|c|}{CrossAligned}                                                 & 17.02    & 4.34    & 8.59    & 74.8                                 & 69.13   & 319.1  & 148.53                                  & {\color[HTML]{FE0000} \textbf{88.12}}     \\ \hline
\multicolumn{1}{|c|}{StyleEmbedding}                                               & 71.8     & 13.65   & 31.3    & {\color[HTML]{FE0000} \textbf{8.9}}  & 121.66  & 379.81 & 214.96                                  & 90.56     \\ \hline
\multicolumn{1}{|c|}{multi\_decoder}                                               & 40.81    & 8.24    & 18.33   & {\color[HTML]{FE0000} \textbf{46.4}} & 201.59  & 642.13 & 359.79                                  & {\color[HTML]{FE0000} \textbf{88.35}}      \\ \hline
\multicolumn{1}{|c|}{TemplateBased}                                                & 48.67    & 12.86   & 25.02   & 79.7                                 & 3258.19 & 375.62 & {\color[HTML]{FE0000} \textbf{1106.28}} & 89.71     \\ \hline
\multicolumn{1}{|c|}{DeleteOnly}                                                   & 33.94    & 9.29    & 17.75   & 84.8                                 & 171.66  & 279.55 & 219.06                                  & 89.28     \\ \hline
\multicolumn{1}{|c|}{DeleteAndRetrieve}                                            & 34.48    & 9.82    & 18.4    & 87.7                                 & 137.04  & 343.75 & 217.04                                  & 89.39     \\ \hline
\multicolumn{1}{|c|}{RetrieveOnly}                                                 & 0.88     & 0.43    & {\color[HTML]{FE0000} \textbf{0.61}}    & 98.4                                 & 150.54  & 150.62 & 150.58                                  & \color[HTML]{FE0000} \textbf{86.33}     \\ \hline
\multicolumn{1}{|c|}{BackTranslation}                                              & 0.67     & 0.52    & {\color[HTML]{FE0000} \textbf{0.59}}    & 96.2                                 & 30.53   & 148.77 & 67.39                                   & \color[HTML]{FE0000} \textbf{87.36}     \\ \hline
\multicolumn{1}{|c|}{UnpairedRL}                                                   & 42.29    & 10.6    & 21.17   & {\color[HTML]{FE0000} \textbf{47.7}} & 328.8   & 735.1  & 491.63                                  & 88.51     \\ \hline
\multicolumn{1}{|c|}{DualRL}                                                       & 58.72    & 17.71   & 32.25   & 86.8                                 & 87.72   & 273.73 & 154.96                                  & 92.14      \\ \hline
\multicolumn{1}{|c|}{B\_GST}                                                       & 43.45    & 13.49   & 24.21   & 86.1                                 & 165.59  & 184.02 & 174.57                                  & 91.78     \\ \hline
\multicolumn{1}{|c|}{G\_GST}                                                       & 43.94    & 13.28   & 24.15   & 77.2                                 & 441.38  & 274.25 & 347.92                                  & 91.15     \\ \hline\hline
\multicolumn{1}{|c|}{\begin{tabular}[c]{@{}c@{}}human: DRG\end{tabular}}       & 26.97    & 53.35   & 37.93   & 72.8                                 & 121.17  & 153.45 & 136.36                                  & 95.83      \\ \hline
\multicolumn{1}{|c|}{human: DualRL}                                                & 36.79    & 33.02   & 34.86   & 77                                   & 178.63  & 196.15 & 187.19                                  & 95.83     \\ \hline
\multicolumn{1}{|c|}{input copy}                                                   & 100      & 21.01   & 45.84   & 3.5  & 69.72   & 131.91 & 95.9                                    & 93.18     \\ \hline
\end{tabular}
}
\caption{Automatic evaluation results of the Yelp dataset (s: self, h: human, G: geometric mean, f: fine-tuned, p: pre-trained). The red indicates that the evaluation score is significantly worse than other systems. Our model is referred to as SST($\alpha, \beta$). The bold black indicates the better performance of our systems for the four metrics that determine it is a stable system.}
\label{Tab:Yelp}
\end{table*}

\begin{table*}[!h]
\centering
\resizebox{2.0\columnwidth}{!}{
\begin{tabular}{c|c|c|c|c|c|c|c|c|}
\cline{2-9}
                                        & \multicolumn{3}{c|}{Content}                           & Attribute                            & \multicolumn{3}{c|}{Fluency}                                & Semantic   \\ \hline
\multicolumn{1}{|c|}{Model}             & s-BLEU & h-BLEU & G-BLEU                               & Classifier($\%$)                           & d-PPL    & g-PPL  & t-PPL                                   & BERTscore  \\ \hline\hline
\multicolumn{1}{|c|}{SST \scriptsize{(0.6, 0.5)}} & 45.47  & 20.34  & \textbf{30.41}                                & \textbf{66.5}                                 & 4.51     & 367.73 & \textbf{40.72}                                   & \textbf{89.17}  \\ \hline\hline
\multicolumn{1}{|c|}{CrossAligned}                                                 & 0.76    & 0.61    & \color[HTML]{FE0000} \textbf{0.68}    & 74.8                                 & 1.11   & 119.37  & 11.51                                  & \color[HTML]{FE0000} \textbf{85.95}     \\ \hline
\multicolumn{1}{|c|}{StyleEmbedding}    & 32.03  & 12.95   & 20.37                                & \color[HTML]{FE0000} \textbf{42.4}                                 & 3.42     & 369.24 & 35.54                                   & 87.39  \\ \hline
\multicolumn{1}{|c|}{multi\_decoder}    & 16.48  & 6.61   & 10.44                                & 70.3                                 & 1.39     & 343.72 & 21.86                                   & \color[HTML]{FE0000} \textbf{86.09}  \\ \hline
\multicolumn{1}{|c|}{TemplateBased}     & 68.54  & 33.79  & 48.12                                & 64.8                                 & 5.36     & 368.41 & 44.44                                   & 90.65 \\ \hline
\multicolumn{1}{|c|}{DeleteOnly}        & 57.48  & 28.56  & 40.52                                & {\color[HTML]{FE0000} \textbf{50}}   & 2.78     & 251.24 & 26.43                                   & 90.55 \\ \hline
\multicolumn{1}{|c|}{DeleteAndRetrieve} & 60.75  & 30.83  & 43.28                                & {\color[HTML]{FE0000} \textbf{52.4}} & 2.43     & 221.92 & 23.22                                   & 90.92  \\ \hline
\multicolumn{1}{|c|}{RetrieveOnly}      & 2.82   & 1.23   & {\color[HTML]{FE0000} \textbf{1.86}} & 82.3                                 & 5.65     & 135.22 & 27.64                                   & \color[HTML]{FE0000} \textbf{85.54}  \\ \hline
\multicolumn{1}{|c|}{B\_GST}            & 58.21  & 25.47  & 38.5                                 & 59.1                                 & 12448.44 & 193.73 & {\color[HTML]{FE0000} \textbf{1552.94}} & 91.23  \\ \hline
\multicolumn{1}{|c|}{G\_GST}            & 51.02  & 21.1   & 32.81                                & 57.3                                 & 18106    & 458.93 & {\color[HTML]{FE0000} \textbf{2882.6}}  & 89.48 \\ \hline\hline
\multicolumn{1}{|c|}{human: DRG}        & 47.67  & 100    & 69.04                                & 46.9                                 & 12.38    & 132.18 & 40.45                                   & 100        \\ \hline
\multicolumn{1}{|c|}{input copy}        & 100    & 47.6   & 68.99                                & 18.5                                 & 3.76     & 188.33 & 26.61                                   & 93.77 \\ \hline
\end{tabular}
}
\caption{Automatic evaluation results of the Amazon dataset. Evaluation metrics are the same as Yelp, but BackTranslation, UnpariedRL, and DualRL do not provide results from Amazon datasets.}
\label{Tab:Amazon}
\end{table*}

\subsection{Human Evaluation}
\begin{table}[!t]
    \centering
    \resizebox{0.85\columnwidth}{!}{
        \begin{tabular}{|c|c|c|c|}
        \hline
        Model           & Content                           & Fluency                           & Style                             \\ \hline
        SST\scriptsize{(0.7, 0.75)}            & 3.32                              & 3.37                              & 3.3                               \\ \hline
        BackTranslation & {\color[HTML]{FE0000} \textbf{2.69}} & {\color[HTML]{FE0000} \textbf{3.15}} & 2.99                              \\ \hline
        TemplatedBased  & 3.18                              & 3.16                              & 3.19                              \\ \hline
        StyleEmbedding  & 3.56                              & 3.49                              & {\color[HTML]{FE0000} \textbf{2.88}} \\ \hline
        \end{tabular}
    }
    \caption{Human evaluation results. The higher the number, the better. Red means the worst result in the corresponding evaluation term.}
    \label{Tab:human}
\end{table}


Table~\ref{Tab:human} shows human evaluation results for content, fluecy, and style. Comparison models, StyleEmbedding and TemplatedBased, each have weaknesses in attribute and fluency. BackTranslation has weaknesses in content and semantic in automatic evaluation. In the yelp test set, we randomly sampled 250 and gave 6 people hired through the Amazon mechanical turk~\footnote{https://www.mturk.com/} evaluate content, fluency, and style between 1 and 5 points. As a result, BackTranslation and StyleEmbedding show the worst results for content, fluency, and style, respectively. Since humans evaluate fluency from a general point of view, the fluency performance of BackTranslation, which is poor in overall performance, and TemplateBased, which has poor t-PPL performance, are similarly bad. We confirm that our system has adequate performance in human evaluation as well as automatic evaluation.

\subsection{Result Analysis}
Human systems do not obtain the highest performance scores except for human-BLEU and BERTscore, which are calculated using human references. \textit{But which of the sentences in human and machines is actually realistic? Probably human.} It is difficult to determine the best system with only automatic evaluation, but it is possible to determine which system is stable or unstable. If a system has significantly lower performance during the evaluation, it is considered unstable. The stable systems in the Yelp dataset are SST, DeleteOnly, DeleteAndRetireve, DualRL, B-GST, and G-GST. For the Amazon dataset, the stable systems are SST and TemplateBased. For all the metrics in both datasets, the stable systems are SST and DualRL. In automatic evaluation, DualRL outperforms SST, but DualRL does not share the model parameters of positive to negative and negative to positive tasks. Therefore, direct comparison is difficult because DualRL is regarded as two models.

We trained SST by changing the random seed of the model initialization several times and found that SST can always yield similar and comparable results. SST can be inferred as a stable system for the following reasons: (1) \textbf{G-BLEU}: Delete and Generate approaches show the stable performance of G-BLEU because the methods generate a sentence based on content tokens. There is no guarantee that content tokens will always be maintained, but content tokens help the generator. (2) \textbf{Attribute}: Our delete process is a method of determining whether certain tokens are deleted with \textit{Important Score}. The direct and model-agnostic deletion is effective for neutralizing sentences. SST also improves a style accuracy by adding style control loss. (3) \textbf{Fluency}: TemplatedBased, B-GST, and G-GST show non-ideal fluency in d-PPL. TemplatedBased is considered unstable because it simply inserts attribute tokens of training data when generating test sentences. Since B-GST and G-GST use pre-trained GPT, they also have the ability to predict the distribution of tokens that are not in training data. The ability to predict generalized tokens is usually helpful, but can sometimes be harmful to d-PPL. SST, the Transformer encoder-decoder structure, learns only the distribution of given data and therefore has a stable d-PPL. (4) \textbf{Semantic}: Transformer language modeling is known to perform better on various tasks than RNN. Even in the style transfer task, the Transformer-based structures seem to reflect the linguistic characteristics.

We observed that unstable systems performed poorly in human evaluation as well in automatic evaluation. However, since performing human evaluation every time is expensive, choosing a stable system with automatic evaluation can be helpful.

Table~\ref{Tab:compare_papers} shows the samples of the generation of the models, which shows the lack of comparison models. In Yelp's negative to positive example, there are only SST and DualRL models that change the style while preserving content that includes \textit{taste} and \textit{price} of the food. In Yelp's positive to negative example, the \textit{professionals} word contains a combination of style and content. In this case, the deletion and generation framework has the disadvantage of corrupting content information.

\begin{table*}[!t]
\centering
\resizebox{2.0\columnwidth}{!}{
\begin{tabular}{|c|c|c|}
\hline
{\color[HTML]{3531FF} \textbf{}}                           & Yelp ({\color[HTML]{FE0000} \textbf{negative}} to {\color[HTML]{3531FF} \textbf{positive}}) & Yelp ({\color[HTML]{3531FF} \textbf{positive}} to {\color[HTML]{FE0000} \textbf{negative}})\\ \hline
\begin{tabular}[c]{@{}c@{}}Input (source)\end{tabular} & the food was so-so and very {\color[HTML]{FE0000} \textbf{over priced}} for what you get .              & these two women are {\color[HTML]{3531FF} \textbf{professionals}} . \\ \hline\hline
SST                                                        & the service is so-so and very {\color[HTML]{3531FF} \textbf{reasonably priced}} for what you get .         & these two women are {\color[HTML]{FE0000} \textbf{rude}} .                       \\ \hline
CrossAligned                                               & the food was {\color[HTML]{3531FF} \textbf{fantastic}} and very very {\color[HTML]{3531FF} \textbf{nice}} for what you .                     & these two dogs are {\color[HTML]{FE0000} \textbf{hard}} down . \\ \hline
StyleEmbedding                                             & the food was so-so and very {\color[HTML]{FE0000} \textbf{over priced}} for what you get .                      & these two pot everywhere was . \\ \hline
DeleteOnly                                                 & the food was so-so and very {\color[HTML]{FE0000} \textbf{over priced}} for what you get . & i {\color[HTML]{FE0000} \textbf{would n't like}} these two women are professionals . \\ \hline
DeleteAndRetrieve          & \begin{tabular}[c]{@{}c@{}}the service is {\color[HTML]{3531FF} \textbf{fantastic}} and the food was so-so \\ and the food is {\color[HTML]{FE0000} \textbf{very priced}} for what you get .\end{tabular}  & these two {\color[HTML]{FE0000} \textbf{scam}} women are {\color[HTML]{3531FF} \textbf{professionals}} . \\ \hline
Back-translation                                           & the food is {\color[HTML]{3531FF} \textbf{delicious}} and the staff are very {\color[HTML]{3531FF} \textbf{good}} for me .                    & this place is just {\color[HTML]{FE0000} \textbf{not good}} .  \\ \hline
UnpariredRL                                                & the food was so-so and very {\color[HTML]{FE0000} \textbf{over priced}} for what {\color[HTML]{3531FF} \textbf{great}} qualities .                & these two women are {\color[HTML]{3531FF} \textbf{great}} .    \\ \hline
DualRL                                                     & the food was {\color[HTML]{3531FF} \textbf{surprising}} and very {\color[HTML]{3531FF} \textbf{reasonably priced}} for what you get .      & these two women are {\color[HTML]{FE0000} \textbf{unprofessional}} .    \\ \hline
B-GST                                                      & the food was {\color[HTML]{3531FF} \textbf{amazing}} - so {\color[HTML]{3531FF} \textbf{fresh}} and very {\color[HTML]{3531FF} \textbf{good}} for what you get .              & these two women are {\color[HTML]{FE0000} \textbf{terrible}} liars .         \\ \hline
G-GST                                                      & the food was {\color[HTML]{3531FF} \textbf{priced right}} - so {\color[HTML]{3531FF} \textbf{nice}} and very {\color[HTML]{3531FF} \textbf{good}} for what you get .                 & these two women are {\color[HTML]{FE0000} \textbf{condescending}} .        \\ \hline
Human\_DRG                                                 & the food was {\color[HTML]{3531FF} \textbf{great}} and {\color[HTML]{3531FF} \textbf{perfectly priced}}      & these two women are {\color[HTML]{FE0000} \textbf{not professionals}} .   \\ \hline
Human\_DualRL                                              & the food was {\color[HTML]{3531FF} \textbf{good}} and the {\color[HTML]{3531FF} \textbf{price is low}} .                  & these two women are {\color[HTML]{FE0000} \textbf{not professionals}} at all      \\ \hline
\hline
                  & Amazon ({\color[HTML]{FE0000} \textbf{negative}} to {\color[HTML]{3531FF} \textbf{positive}}) & Amazon ({\color[HTML]{3531FF} \textbf{positive}} to {\color[HTML]{FE0000} \textbf{negative}})    \\ \hline
Input (source)    & i have to {\color[HTML]{FE0000} \textbf{lower}} the rating another notch .                    & it seems to be of very {\color[HTML]{3531FF} \textbf{good}} quality in its build .                                                                                    \\ \hline\hline
SST               & {\color[HTML]{3531FF} \textbf{love}} the rating another one , & it seems to be of very {\color[HTML]{FE0000} \textbf{poor}} quality in its build .                                                                                    \\ \hline
CrossAligned      & i would {\color[HTML]{3531FF} \textbf{recommend}} this for the price .                      & it s {\color[HTML]{FE0000} \textbf{not be for a good}} game for my phone .                                                                                            \\ \hline
StyleEmbedding    & i have to {\color[HTML]{3531FF} \textbf{get}} by a one market .                  & it seems to be the num\_extend is {\color[HTML]{3531FF} \textbf{good nice}} high cases .                                                                              \\ \hline
DeleteOnly        & i have to {\color[HTML]{FE0000} \textbf{lower}} the rating and it fits into another notch .                     & \begin{tabular}[c]{@{}c@{}}i have previously charged num\_num \\ different bt headsets that last num\_num hours longer .\end{tabular} \\ \hline
DeleteAndRetrieve & i have to {\color[HTML]{FE0000} \textbf{lower}} the rating another notch and i {\color[HTML]{3531FF} \textbf{love}} it .      & initially it was very {\color[HTML]{3531FF} \textbf{good}} quality in its build .                                                                                     \\ \hline
B-GST             & i have {\color[HTML]{FE0000} \textbf{lower}} levels for the other notch .                   & it seems to be of very {\color[HTML]{3531FF} \textbf{good}} quality in taste .                                                                                        \\ \hline
G-GST             & i have {\color[HTML]{FE0000} \textbf{lower}} the steel another notch .     & it seems to be of very {\color[HTML]{3531FF} \textbf{good}} value in return .                                                                                         \\ \hline
Human\_DRG        & i have to {\color[HTML]{3531FF} \textbf{raise}} the rating another notch .                                             & it seems to be of very {\color[HTML]{FE0000} \textbf{poor}} quality in its build                                                                                      \\ \hline
\end{tabular}
}
\caption{Examples of comparison of generated sentences of AI systems. SST is our model. Attributes are colored. Red is negative and blue is positive.}
\label{Tab:compare_papers}
\end{table*}

\subsection{Ablation Study}
If we use style loss for SST training, Table~\ref{Tab:ablation} shows that the style accuracy has 4 point gain. Fluency and semantic are slightly better. It is observed that style loss improves the data-fluency, resulting in better total fluency. However, style loss decreases G-BLUE slightly by allowing the transferred sentence to change the attribute better.

\subsection{Trade-off between Content and Style}
\label{sec:trade-off}
\begin{table}[!t]
\centering
\resizebox{\columnwidth}{!}{
\begin{tabular}{|c|c|c|c|c|}
\hline
               & Con & Attr  & Flu & Sem  \\ \hline
Model          & G-BLEU  & Cls($\%$) & t-PPL   & BERTscore \\ \hline\hline
SST \scriptsize{(0.7, 0)} & 19.11   & 82.2       & 306.65  & 89.96     \\ \hline
- Style loss   & 19.78   & 78.2       & 341.51  & 89.84     \\ \hline
\end{tabular}
}
\caption{Ablation result of style loss in the Yelp dataset. (Con: content, Attr: attribute, Flu: Fluency, Sem: Semantic)}
\label{Tab:ablation}
\end{table}

\begin{figure}[!t]
\centering
\includegraphics[width=.44\columnwidth]{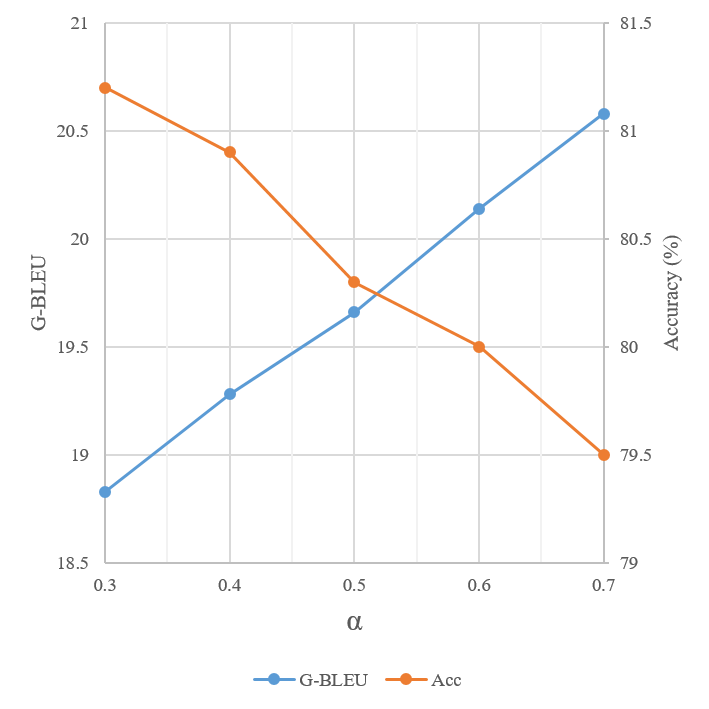}
\includegraphics[width=.44\columnwidth]{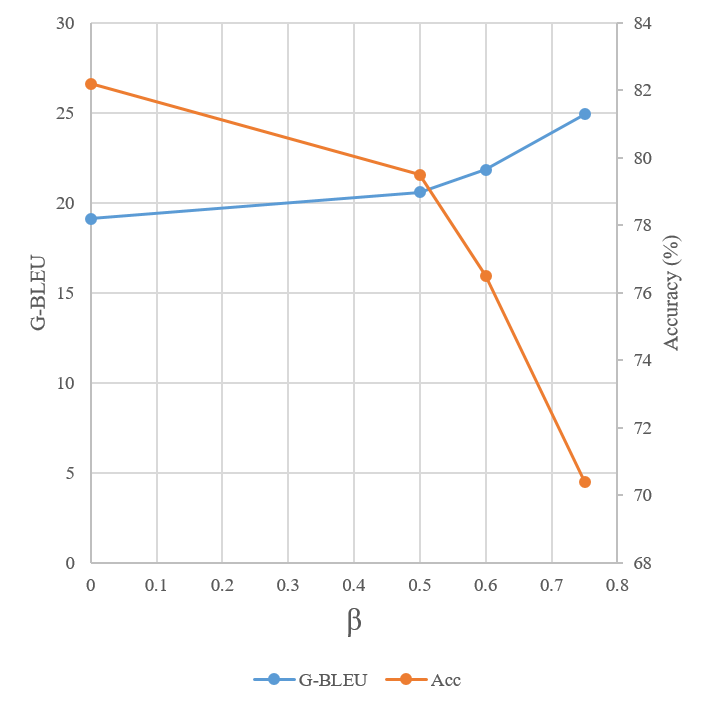}
\label{fig:beta}

\caption{
(a) Trade-off curve of G-BLEU and Style accuracy according to $\alpha$ (at $\beta$ = 0.5) in Yelp
(b) Trade-off curve of G-BLEU and Style accuracy according to $\beta$ (at $\alpha$ = 0.7) in Yelp.
}
\label{fig:trade}
\end{figure}

With $\alpha$ and $\beta$ we can simply adjust the trade-off of content and style. The results of Yelp are shown in Fig.~\ref{fig:trade}. Smaller $\alpha$ and $\beta$ allow the model to focus on style changes, while larger $\alpha$ and $\beta$ allow the model to focus on content preserving. The trade-off of content and style changes linearly with $\alpha$ and is sensitive to $\beta$. The appropriate $\alpha$ and $\beta$ depend on datasets.

\subsection{Latent Space Walking}
\begin{table}[!t]
\centering
\resizebox{\columnwidth}{!}{
\begin{tabular}{|c|c|}
\hline
Source(negative) & \begin{tabular}[c]{@{}c@{}}when i was finally there , i was very disappointed .\end{tabular} \\ \hline
after deletion     & \begin{tabular}[c]{@{}c@{}}when i finally , i very .\end{tabular}                            \\ \hline\hline
style: negative    & \begin{tabular}[c]{@{}c@{}}when i finally left , i was very disappointed .\end{tabular}      \\ \hline
\multirow{3}{*}{↓} & \begin{tabular}[c]{@{}c@{}}when i finally left , i was very disappointed.\end{tabular}      \\ \cline{2-2} 
                  & when i finally walked in , i was very disappointed .\\ \cline{2-2} 
                  & when i finally got , i was very happy . \\ \hline
style: positive    & \begin{tabular}[c]{@{}c@{}}when i finally got , i was very happy .\end{tabular}              \\ \hline
\end{tabular}
}
\caption{One sample of the Yelp dataset. SST generates a sentence from style vector space to negative to positive}
\label{Tab:latent}
\end{table}

In this section we observe the transferred sentences according to the weight of positive and negative in the continuous style vector space. Ideally, a neutral sentence should be generated when the style attribute has the same weight for negative and positive. An example is shown in Table~\ref{Tab:latent}. A lot of data, like this example, don't show a neutral sentence even if the style has the same weight for the negative and positive. If we train our model to reflect this problem, we can expect better style control.

\section{Conclusion and Future Work}
We propose Stable Style Transformer (SST) that re-writes the sentences with Delete and Generate. SST is a system that can be used in the real world with overall stable results compared to other comparable systems. We show that filtering out unstable systems through human evaluation is expensive, so selecting a stable system through automatic evaluation can be helpful. The proposed direct and model-agnostic deletion method allows the classifier to intuitively delete attribute markers and easily handle the trade-off of content and style. In future work, we will study solutions for the case where attribute markers also contain content in the deletion and generation framework.

\bibliography{anthology,acl2020}
\bibliographystyle{acl_natbib}
\end{document}